\documentclass[letterpaper]{article} 
\usepackage{aaai21}  
\usepackage{times}  
\usepackage{helvet} 
\usepackage{courier}  
\usepackage[hyphens]{url}  
\usepackage{graphicx} 
\usepackage{natbib}  
\usepackage{caption} 
\usepackage{latexsym}
\usepackage{amssymb}
\usepackage{amsmath}
\usepackage{graphicx}
\usepackage{url}
\usepackage{booktabs}
\newcommand{\ra}[1]{\renewcommand{\arraystretch}{#1}}
\usepackage{pifont}
\usepackage{multirow}
\usepackage{xcolor}

\title{Answering Complex Queries in Knowledge Graphs with Bidirectional Sequence Encoders}
\author{
    Bhushan Kotnis \& 
Carolin Lawrence \&  
 Mathias Niepert\\}
   \affiliations{
	NEC Laboratories Europe GmbH, \\
	Heidelberg, Germany 69115\\
	$\left\{bhushan.kotnis, carolin.lawrence, mathias.niepert\right\}$@neclab.eu}

\begin{document}

\maketitle
\begin{abstract}
Representation learning for knowledge graphs (KGs) has focused on the problem of answering simple link prediction queries. In this work we address the more ambitious challenge
of predicting the answers of conjunctive queries with multiple missing entities. We propose Bidirectional Query Embedding (\textsc{BiQE}), a method that embeds conjunctive queries with models based on bi-directional attention mechanisms. Contrary to prior work, bidirectional self-attention can capture interactions among all the elements of a query graph. We introduce two new challenging datasets for studying conjunctive query inference and conduct experiments on several benchmark datasets that demonstrate \textsc{BiQE} significantly outperforms state of the art baselines.
\end{abstract}

\section{Introduction}\label{sec:introduction}
Linked data structures such as graphs and specifically Knowledge Graphs (KG) are well suited for representing a wide variety of heterogeneous data. Knowledge graphs represent real-world entities along with their types, attributes, and relationships. 
Most existing work on machine learning for knowledge graphs has focused on \emph{simple} link prediction problems where the query asks for a single missing entity (or relation type) in a single triple. 
A major benefit of knowledge graph systems, however, is their support of a wide variety of logical queries. For instance, \textsc{SPARQL} a typical query language for RDF-based knowledge graphs supports a variety of query types. However, it can query only for facts that exist in the database, it cannot infer missing knowledge. 

\begin{figure*}[t!]
	\centering
	\includegraphics[width=0.98\textwidth]{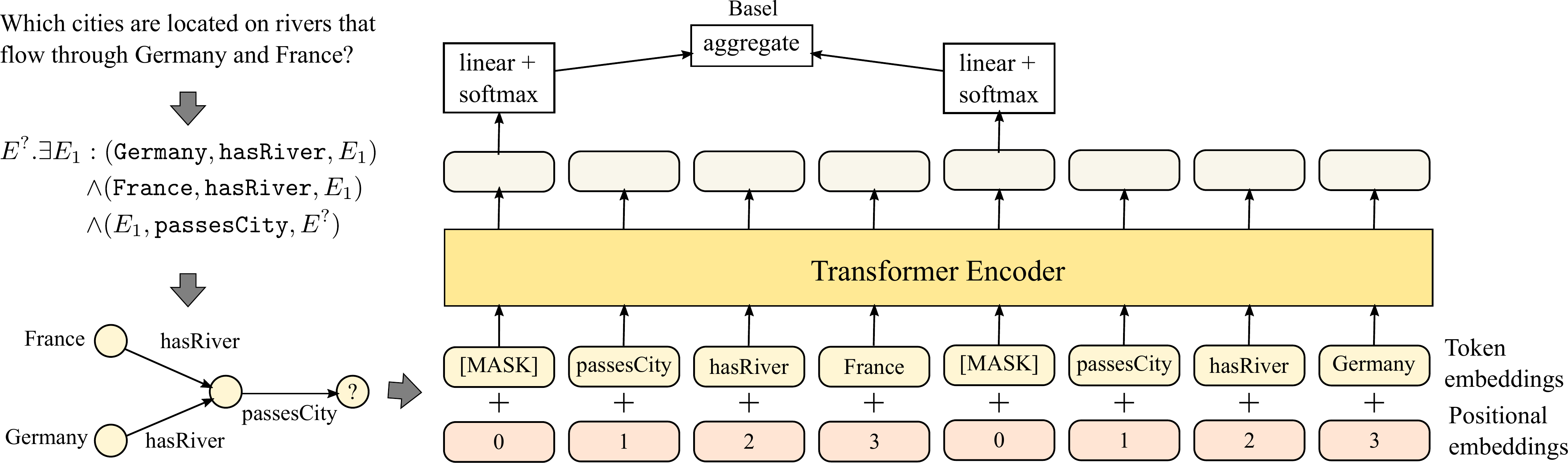}
	\caption{\textsc{BiQE} model architecture and workflow.}
	\label{fig:model}
\end{figure*}

To address this shortcoming, we are interested in the problem of computing probabilistic answers to \emph{conjunctive queries} (see for example Figure~\ref{fig:model}) that can be mapped to subgraph matching problems and which form a subset of \textsc{SPARQL}. Every query can be represented with a graph pattern, which we refer to as the \emph{query graph}, with some of its entities missing.
Path queries are conjunctive queries that can be expressed using a linear recursion, while directed acyclic graph queries express at least one binary recursion. For instance, the query illustrated in Fig. \ref{fig:model} 
\textit{"Name cities that are located on rivers that flow through Germany and France?"}  is an instance of such a conjunctive query (see Fig. \ref{fig:model}). We address answering conjunctive query composed from entities and relations of an incomplete KG.

Note that this is not a question answering (KBQA) task \cite{Hu2018} nor a subtask of KBQA. In KBQA the knowledge graph is assumed to be complete and the task is to match the natural language query to an appropriate subgraph. In our setting, a conjunctive query is provided. 

Prior work proposed methods for answering path queries by composing standard scoring functions such as \textsc{TransE}~ \cite{Bordes2013} and \textsc{DistMult}~\cite{Yang2014} used in knowledge graph completion models for link prediction~\cite{Guu2015}. Models that encode sequences such as RNNs,\cite{Das2017L}, were also used for encoding paths. 
More recent methods addressed the problem of answering queries composed of several paths intersecting at a missing entity, e.g., ~\cite{Hamilton2018, Ren2020}, by composing triple scoring functions along with neural network based intersection operators. Building on this, \cite{Daza2020} propose relational Graph Convolutional Networks for answering conjunctive queries. 

The problem with recent methods which use intersection operators or GCNs is that such models aggregate information only along the paths starting from the source and ending at the intersection (see Fig. \ref{fig:embedding} (left)). These approaches, however, cannot model complex dependencies between various parts of the query that are not directly connected. Additionally such models cannot jointly infer missing entities in queries with more than one missing entity.

We propose to explicitly model more complex dependencies between the components of the queries and to answer queries that are not restricted to a single missing entity. Answering these queries, however, is not straightforward. For instance, in the example of Fig. \ref{fig:model}, one needs to consider all rivers that flow through both countries and perform a set intersection for obtaining the common rivers. 

To address the challenge of answering novel query types and the shortcomings of existing approaches, we propose \textsc{BiQE}, a \textit{Bi-directional Query Encoder}, that uses a bidirectional transformer to incorporate the entire query context. 
There is no obvious way to feed a query graph to a transformer~\cite{Vaswani2017} because sequences contain positional information while the various branches of a query graph are permutation invariant. In this paper, we propose a novel positional encoding scheme that allows a transformer to answer conjunctive graph queries. 
We specifically focus on answering conjunctive queries that (a) can be represented as directed acyclic graphs (DAGs) and (b) have mutliple query variables, as this problem has received little attention.




\begin{itemize}
	\item A method for jointly predicting the answers of queries with more than one missing entity (query variable);
	\item An elegant approach for encoding query DAGs into bidirectional transformer models; 
	\item Two new benchmarks for complex query answering in KGs where queries contain multiple missing entities;
	\item Empirical results demonstrating that our method outperforms the state of the art;
	\item Extensive qualitative analysis of the behavior of the attention mechanism.
\end{itemize}

\section{Problem Statement}\label{sec:task}

A knowledge graph $\mathcal{G} = (\mathcal{E}, \mathcal{R}, \mathcal{T})$ consists of a set of entities $\mathcal{E}$, a set of relation types $\mathcal{R}$, and a set of triples $\mathcal{T}$ of the form $t = (e,r,e') \in \mathcal{E} \times \mathcal{R} \times \mathcal{E}$. 
Let us denote $E_1, ..., E_k$ as existentially quantified variables, $E^{?}_1, ..., E^{?}_k$ as free variables, and $e_i$ as some arbitrary entity. The free variables represent the entities to be predicted.
We specifically focus on conjunctive queries of the form
\begin{align}
	\label{eqn:query}
	E^{?}_{1}, ..., E^{?}_{k} . \exists E_1,... , E_{m} : c_1 \wedge c_2 \wedge ... \wedge c_n,
\end{align}
where $c_{\ell}, \ell \in \{1, ..., n\}$ 
is one of the following 
\begin{enumerate} 
	\setlength\itemsep{0.08mm}
	\item $(e_i,r,E^{?}_{j})$  or $(E^{?}_{j},r,e_i)$;
	\item $(e_i,r,E_{j})$ or $(E_{j},r,e_i)$;
	\item $(E_{i},r,E_{j})$ or $(E^{?}_{i},r,E^{?}_{j})$; or
	\item $(E_{i},r,E^{?}_{j})$ or $(E^{?}_{i},r,E_{j}).$
\end{enumerate}

The \emph{query graph} for a conjunctive query is the graph consisting of triples $c_1\dots c_n$ of equation (\ref{eqn:query}) and of the types (1)-(4). We constrain the set of conjunctive queries to those for which the query graph is a connected directed acyclic graph (DAG). The \emph{depth} of a vertex in a DAG is the maximal length of a path from a root node to this vertex. We require that any two missing entity nodes have different depth\footnote{The depth constraint is required since two different query variables at the same depth would be assigned the same positional encoding.}.
This is similar to the definition of conjunctive queries in previous work~\cite{Hamilton2018} with the exception that we can have more than one free variable in the query and that free variables can be at various positions in the DAG. A typical query graph is illustrated in the left top half of Figure~\ref{fig:model}.

\section{The Bidirectional Query Encoder}

We aim to model interactions between all the elements in a query structure such that the model supports joint inference of more than one missing entities. For this purpose we use a bi-directional transformer \citep{bert:2018} to encode conjunctive graph queries. One of the  crucial features of the transformer is its self-attention module which allows every token given to the model to simultaneously attend to every other token.
A transformer consists of several layers, the first one being the embedding layer that sums up the token and positional embedding for each token in the sequence. The positional embeddings receive the position of a token in the sequence which ranges from $0$ to the maximum sequence length. The embedding layer is followed by several encoder layers, each containing a self-attention layer followed by a feed forward layer followed by layer normalization. 
It has been observed that the self-attention mechanism acts like a fully-connected graph neural network because it induces a weighted complete (latent) graph on the set of input tokens. In the context of complex query answering, this  allows our proposed model to induce a latent dependency structure between query tokens in addition to the observed one.  

The query DAG corresponding to a conjunctive query under consideration can have multiple nodes representing free variables $E^?$ (target nodes; limited to one per depth), bound variables $E$ (quantifier nodes), and entities $e$ (anchor nodes). The input for Transformer models is a sequence of tokens that has a naturally defined total order. DAGs differ from sequences in two important aspects. First, nodes in graphs can have multiple predecessors and successors while tokens in a sequence have only one predecessor and one successor. Second, nodes in DAGs are not totally but only partially ordered. As a consequence, using DAGs as input for Transformer models is not straight-forward. We address this challenge by decomposing the query DAG into a set of query paths from each root node to each leaf node of the query DAG. The DAG structure imposes a partial ordering on the nodes which allows us to decompose the query DAG into a set of path queries that originate from root nodes and end in leaf nodes. Any query DAG that is a tree with $m$ root and $n$ leaf nodes can be decomposed into $mn$ paths. For general query DAGs the number of paths can be exponential in the number of nodes in the worst case. This is not a problem in practice, however, as the size of conjunctive queries is typically small and assumed fixed. Indeed, the \emph{combined} query complexity of conjunctive queries, that is, the complexity with respect to the size of the data and the query itself, is NP-hard in relational databases. We can therefore not expect a query answering algorithm for the more challenging problem of conjunctive queries with missing data to be less complex. 

Since there is an order within each path but no ordering between paths, we use \emph{positional encodings} to represent the order of paths. The paths are fed to the transformer encoder tail first with positional id $0$ for the $[\mathtt{MASK}]$ token representing the tail query. The positional ids are reset to $0$ at the position at every path boundary. An example is depicted in Fig.~\ref{fig:model}.  Because the self-attention layers are position invariant and the positional information lies solely in the position embeddings, the positional encoding of tokens in a branch does not change even if the order between the branches is changed. This allows us to feed a set of path queries to the transformer in any arbitrary order and allows the model to generalize even when the order of path queries is changed. This is depicted in the right half of Figure~\ref{fig:model}.

\begin{figure*}[t!]
	\centering
	\includegraphics[width=0.96\textwidth]{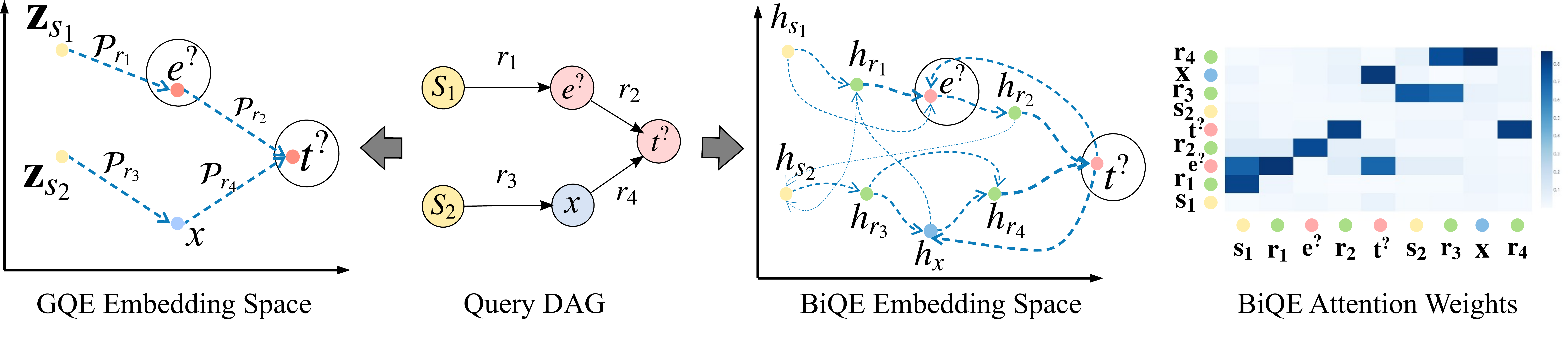}
	\caption{Query embedding in GQE (left) vs. \textsc{BiQE} (right). For GQE, when computing the intersection of \textbf{e} or \textbf{t}, only the previous query context is considered and not the future. In contrast, for \textsc{BiQE}, every element can attend to every other element of the query. This joint modeling of elements leads to a higher accuracy.}
	\label{fig:embedding}
\end{figure*}

We map a single path query to a token sequence by representing free variables (values to be predicted) with $[\mathtt{MASK}]$ tokens and dropping existentially quantified variables. These existentially quantified variables are intermediate entities. We drop existentially quantified variables when translating the query to the sequence of tokens. For instance, the path query 
\[ E^{?}_{1}, E^{?}_{2}.\exists E_1 : (e,r_1,E_1), (E_1,r_2,E^{?}_{1}), (E^{?}_{1},r_3, E^{?}_{2}) 
\]
is mapped to the sequence
\[
[\mathtt{MASK}] \leftarrow r_3  \leftarrow [\mathtt{MASK}] \leftarrow r_2 \leftarrow r_1 \leftarrow e
\]

We remove the existentially quantified variable nodes since the model can learn discriminating information such as the type of these nodes from the adjacent relation types. The query DAG can also contain query variables in non-terminal (leaf) nodes. The above example of a single path illustrates this. In general \textsc{BiQE} can answer queries without constants but because we believe that such queries are not realistic we do not evaluate this in the experiments.

We train the model to predict the entity from the set of all entities at the location of the $[\mathtt{MASK}]$ tokens using a categorical cross-entropy loss. The model architecture is illustrated in Fig. \ref{fig:model}.  Entity and relations are separate tokens in the form of unique identifiers. Since we decompose query DAGs into multiple paths, there might be two or more $[\mathtt{MASK}]$ tokens that we know refer to the same unknown entity. We address this by aggregating the output probability distributions of these $[\mathtt{MASK}]$ tokens during test time. In our experiments, we used the average aggregation operation. 

It is possible to verify that the proposed model is \emph{equivariant} with respect to permutations of the paths of the query DAG. Let $p_1, ..., p_k$ be the $k$ paths corresponding to an arbitrary query DAG and let $\pi_\mathtt{p}$ be a permutation acting on the paths. The permutation $\pi_\mathtt{p}$ induces a permutation $\pi_{\mathtt{t}}$ on the input and positional tokens. It is straight-forward to verify that the model is equivariant with respect to $\pi_{\mathtt{t}}$, that is, when we permute the input and positional tokens with $\pi_{\mathtt{t}}$ and the output tokens with $\pi_{\mathtt{t}}^{-1}$ the model behaves the same way as a model whose input nor output is permuted. Hence, an inductive bias of the model is it being equivariant with respect to permutations of the paths of the query DAG.

\section{Experiments}\label{sec:experiments}

First, we describe the existing state of the art methods proposed for logical query answering in knowledge graphs. Second, we introduce new benchmarks for complex query evaluation and describe their creation and properties. Third, we compare existing methods with \textsc{BiQE} both on the new datasets as well as on an existing benchmarks. We also show that complex query answering for DAG queries is more challenging than for path queries.  Fourth, we analyze the results and the difficulty of complex query types in relation to where the missing entities are located. Fifth, through qualitative analysis, we analyse the ability of \textsc{BiQE} to attend to different parts of the query for query answering. We use the standard BERT architecture as defined in \cite{bert:2018}.

\subsection{Reference Models}
\label{sec:reference}
We use several reference models including recent ones uploaded to pre-print servers. The Graph Query Embedding (GQE) model \citep{Hamilton2018} was one of the first models to address DAG query answering. The GQE model consists of two parts, the projection function and the intersection function. The projection function computes the path query embedding and the intersection function, using a feed forward neural network, computes the query intersection in embedding space. The projection function is a triple scoring function such as TransE \cite{Bordes2013} or DistMult \cite{Yang2014}. A path representation is obtained by applying a scoring function recursively until the target entity is reached ~\cite{Guu2015}. \cite{Hamilton2018} extend the path compositional model of~\cite{Guu2015}  to DAGs by using a feed forward neural network to model the intersection operation. For paths, the GQE model is identical to the path compositional model.

The Query2Box model (Q2B), \cite{Ren2020}, builds upon the GQE model and embeds output of a link prediction query as a box rather than point in a high dimensional vector space. The entity is then modeled as a point inside the box. This allows Q2B to model intersection of sets of entities as intersection of boxes in vector space. This is done using the intersection operator that computes attention over the box centers and then shrinks the box sizes. 

The Message Passing Query Embedding model (MPQE) uses an $l$ layer R-GCN, \cite{Schlichktrull2018}, followed by an aggregation function to obtain query embedding. This query embedding is then used to find the answer using nearest neighbor search.

\cite{Friedman2020} introduce \textsc{TractOR}, a probabilistic mixture model similar to DistMult, but capable of probabilistic interpretation. \textsc{TractOR} can address a variety of logical queries including disjunctive queries in linear time. Although, \textsc{TractOR} does not contain additional parameters for the intersection operator, or attention mechanism like in \textsc{BiQE} and others like GQE, MPQE and  Q2B, the comparison is done for sake of completion.

Following \cite{Ren2020} we compare \textsc{BiQE} against the GQE-Double model. GQE-Double model, introduced in \cite{Ren2020}, is a GQE model with double the embedding dimension compared to GQE. \cite{Ren2020} fix GQE and Q2B embedding dimension to 400 dimensions.

The GQE model is difficult to implement for datasets with arbitrarily shaped DAGs containing multiple missing entities. The GQE implementation provided with the paper only works for 7 types of DAG queries found in the \textsc{Bio} dataset. Therefore, we re-implemented the GQE model~\cite{Hamilton2018} for experiments with the FB15K-237-CQ and WN18RR-CQ datasets. Due to the difficulty in implementing batched training of the GQE intersection function on arbitrary DAG queries, we only use the projection function described in GQE and use mean pooling as the intersection function. To differentiate this from the original GQE model, we term this GQE-MP (GQE with mean pooling).

All of the above models aggregate information about entities and relations that are part of its ancestors or neighborhood. In contrast, \textsc{BiQE} can aggregate information from any part of the query and jointly predict multiple missing entities. 

\subsection{Datasets}\label{sec:dataset}

\begin{figure*}
	\centering
	\includegraphics[width=0.7\textwidth]{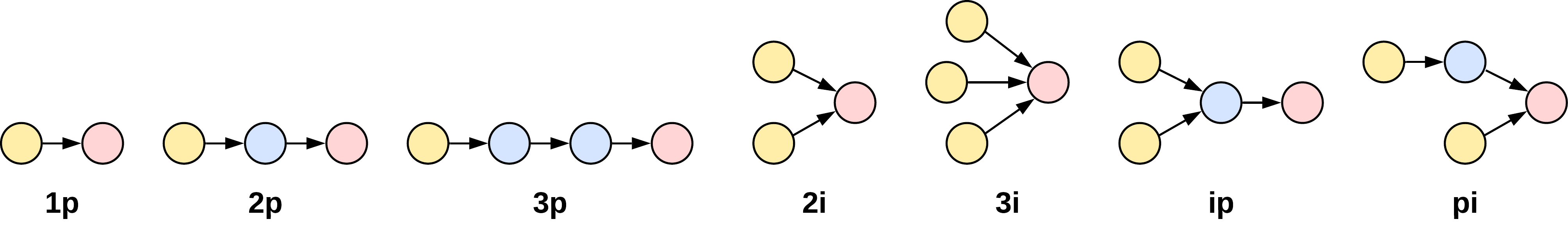}
	\caption{Conjunctive query types present in \textsc{Bio}, FB15K-237 and NELL-995 datasets.}
	\label{fig:queries}
\end{figure*}

\cite{Hamilton2018} introduced two datasets for evaluating conjunctive queries. Of those we use the publicly available \textsc{Bio} dataset.\footnote{The Reddit dataset is not publicly available.} It consists of seven types of conjunctive queries shown in Fig~\ref{fig:queries}. 
The dataset only considers conjunctive queries with exactly one missing entity.

A recent study, \cite{Ren2020}, builds on the work of \cite{Hamilton2018} and introduces three new datasets for evaluating logical queries. These datsets are constructed from FB15K, FB15K-237 and NELL-995 which are publicly available knowledge graph completion datasets. We also evaluate the proposed model on two of these datsets, namely, FB15K-237 and NELL-995. These datasets contain nine types of queries: seven are conjunctive queries identical to those introduced in the \textsc{Bio} dataset. The other two are disjunctive queries. In this work we focus on conjunctive queries and, therefore, we compare \textsc{BiQE} with other models only on conjunctive queries. These queries are illustrated in Fig. \ref{fig:queries}. 


We are interested in studying conjunctive queries with multiple missing entities and none of the existing datasets contain such queries. To address these shortcomings, we introduce two new challenging datasets based on popular KG completion benchmarks, namely FB15K-237 \cite{Toutanova2015_inv} and WN18RR \cite{Dettmers2018}.

Following \cite{Guu2015}, who mine paths using random walks, we sample paths by performing one random walk per node with depth chosen uniformly at random with maximum depth five and then construct DAGs by intersecting the mined random walks. We allow only one intersection point in a DAG and a single target leaf node. We term the datasets containing DAG queries as \textsc{FB15k-237-CQ} and \textsc{WN18RR-CQ}  where CQ represents Conjunctive Queries. To study model performance on DAG queries in comparison to path queries, we create datasets from the mined paths and term them as \textsc{FB15k-237-Paths} and \textsc{WN18RR-Paths}.

We describe the dataset generation process in more detail now. First we mine paths by performing random walks starting from every node in the graph. Note that paths are mined from the appropriate knowledge graph split provided in the original datasets. For each node we: 
\begin{enumerate}
	\item Randomly sample the path depth, that is, a number between 2 and 5 (inclusive).
	\item Select a neighbor at random and obtain the relation type linking the node and neighbor.
	\item Continue with step (2) until the chosen depth is reached. 
\end{enumerate}
We limit the number of mined paths from FB15K-237 train split to 50,000 and WN18RR to 10,000. We term the paths mined from train, validation and test splits as the '\textit{Paths}' dataset, i.e. \textsc{FB15k-237-Paths} and \textsc{WN18RR-Paths}. 

From the generated paths we obtain DAG queries using the following procedure:
\begin{enumerate}
	\item We first intersect the mined paths at intermediate entity positions. This results in sets of paths that have at least one common intermediate entity. The maximum number of intersecting paths is limited to three.
	\item The intersecting entity is made the terminal entity by eliminating the portion of paths succeeding the intersection. The number of intersecting branches is capped at three. This results in a star shaped DAG.
	\item For each star shaped DAG we randomly select an edge incident to the terminal entity. This forms the tail query.   
\end{enumerate}

\begin{figure}[t!]
	\centering
	\includegraphics[width=0.4\textwidth]{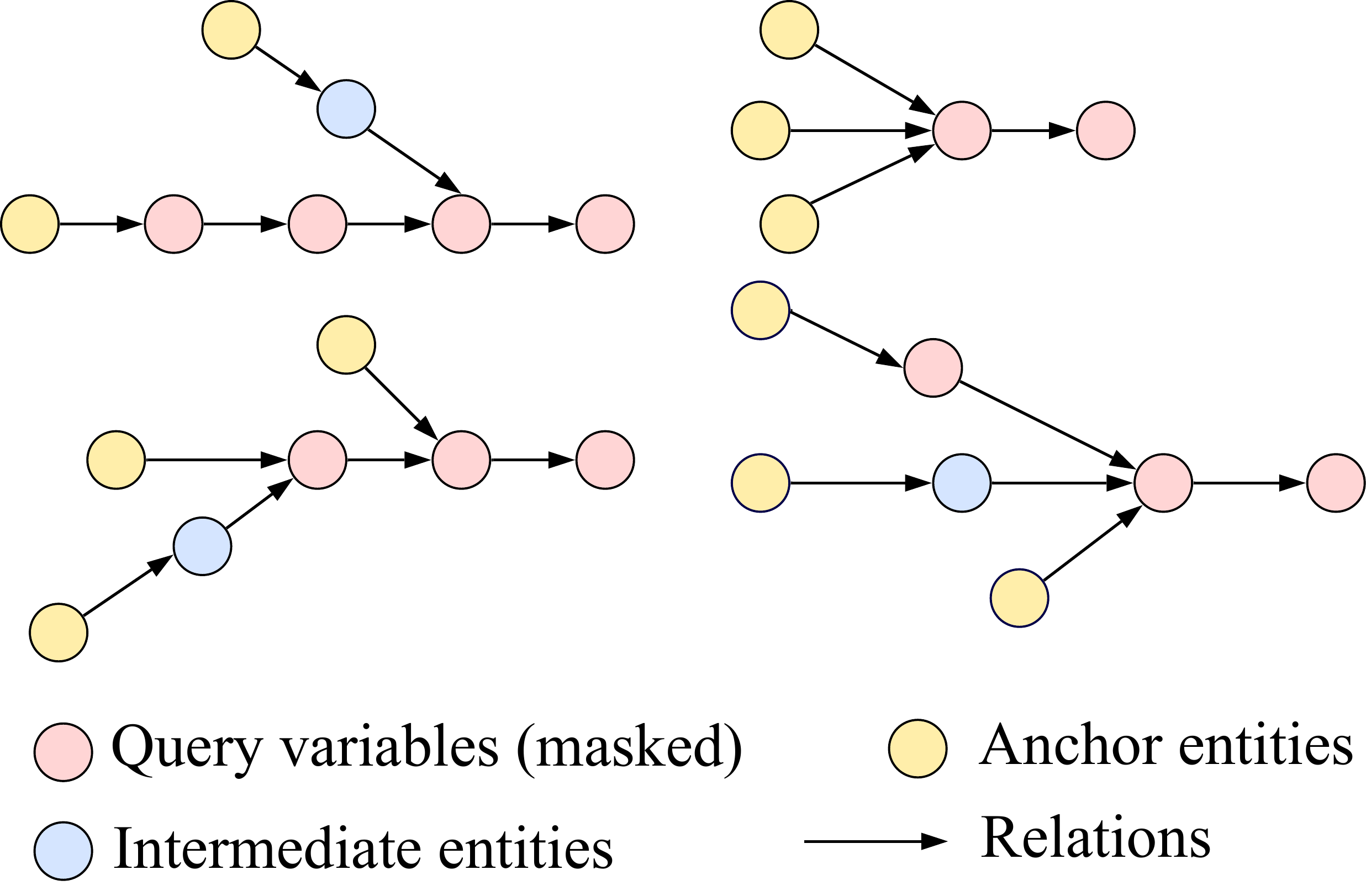}
	\caption{Examples of the random DAG and path queries generated using the procedure described in Section Datasets.}
	\label{fig:dags}
\end{figure}

This procedure results in DAG queries with one target leaf node (tail query), an intersection query and branches with a random number of path lengths. We repeat this for each split of the original dataset. For the development (validation) and test split we remove (mask) the tail entity, the intersection entity, and intermediate entities along the branch.
We provide a few examples of such DAG queries in Fig. \ref{fig:dags}. Note that the random generation procedure using path intersections provides a broader sets of DAG queries compared to the ones illustrated in Fig. \ref{fig:queries}   
We use the generated paths and triples as the training set for \textsc{BiQE} as well as the reference model. This is achieved by predicting the tail entity in a triple and predicting all entities except the source entity in paths during training.

For the CQ datasets, the test and validation splits only contain DAGs and the training split contain all three, that is, triples, paths, and DAGs. Table \ref{table:dataset} describes the dataset statistics for FB15K-237-CQ and WN18RR-CQ. For the Paths dataset, the test and validation splits only contain paths while the training contains paths and triples. The queries present inside test and validation splits in the path dataset contain paths of various lengths ranging from 2 to 5 (inclusive). The task is to predict intermediate entities along with the tail entity in the path, i.e., \emph{all entities except the source entity}. 

Note that the DAG's present in the CQ dataset validation and test splits are obtained by intersecting the paths present in the validation and test splits of the path datasets. This allows us to compare performance of \textsc{BiQE} on paths as well as DAGs that are synthesized from the same paths.

Due to a smaller number of triples and a higher degree of sparsity in WN18RR, we could only obtain about 100 DAGs for the test and validation splits. Therefore, we limited the train DAGs for WN18RR to 10,000.


\begin{table}[!t]
	\centering
	\ra{1.1}
	\small{
		\begin{tabular}{@{}l@{}cccccc@{}}
			\toprule
			& \multicolumn{3}{c}{\textbf{FB15K-237-CQ}}       & \multicolumn{3}{c}{\textbf{WN18-RR-CQ}}        \\
			\cmidrule(lr){2-4} \cmidrule(lr){5-7} 
			& \textbf{train} & \textbf{dev} & \textbf{test} & \textbf{train} & \textbf{dev} & \textbf{test} \\
			\cmidrule(lr){2-2} \cmidrule(lr){3-3} \cmidrule(lr){4-4} \cmidrule(lr){5-5} \cmidrule(lr){6-6} \cmidrule(lr){7-7}\\
			\textbf{Triples} & 272,115 & -    & -    & 86,835 & -   & -  \\
			\textbf{Paths}   & 50,000  & -    &      & 10,000 & -   & -  \\
			\textbf{DAGs}    & 48,865  & 2785 & 2599 & 9465   & 112 & 95 \\
			\textbf{Avg. Masks}    & 1.86   & 5.91 & 6.05 & 1.84   & 5.13 & 4.91 \\
			\textbf{Avg. Len}    & 152   & 460 & 479 & 71   & 198 & 199
			\\ \bottomrule
		\end{tabular}
	}
	\caption{Dataset Statistics. Avg. Len is the average length of sequences feed to \textsc{BiQE}. Average masks and input length include paths and triples.}
	\label{table:dataset}
\end{table}

\subsection{Evaluation Metrics}

For the FB15k-237-CQ path dataset, we evaluate entity prediction using two ranking metrics, namely Mean Reciprocal Rank (MRR) and HITS@10. Entities are ranked based on the scores outputted by the model. MRR is the dataset wide mean of the reciprocal of the rank of the correct entity while HITS@K is the proportion of correct entities that are ranked in the top K ranks. We rank each of the predicted missing query entities against the entire entity set filtering out all positive target entities present in test, validation and train splits. The correct answer to the query is ranked against the filtered entities. This is the filtered setting introduced in \cite{Bordes2013}. For obtaining the filters we performed DAG traversals on the complete graph to obtain a list of positives to be filtered at test time.

We use the same evaluation procedure and code provided with \cite{Ren2020} for evaluating \textsc{BiQE} on FB15K-237 and NELL-995 dataset from \cite{Ren2020}. They use a similar methodology for evaluating HITS@3. For every query, they evaluate HITS@3 for all possible correct answers using the filtered setting and average the scores per query. Using the same evaluation code allows us to compare our model directly with numbers reported in the paper.

For the \textsc{Bio} dataset we follow the evaluation protocols in \cite{Hamilton2018}, i.e., Area Under the Curve (AUC) and Average Percentile Rank (APR). This allows us to compare our model directly with numbers reported in the paper. Due to space constraints we moved the training details and hyperparameter tuning to the appendix which can be found in \cite{Kotnis2020}

\begin{table*}[]
	\centering
	\ra{1.1}
	\small{
		\begin{tabular}{@{}lcccccccc@{}}
			\toprule
			&
			\multicolumn{2}{c}{\textbf{GQE}} &
			\multicolumn{2}{c}{\textbf{GQE-Double}} &
			\multicolumn{2}{c}{\textbf{Q2B}} &
			\multicolumn{2}{c}{\textbf{BiQE}} \\
			\cmidrule(lr){2-3} \cmidrule(lr){4-5} \cmidrule(lr){6-7} \cmidrule(lr){8-9}
			&
			\multicolumn{1}{l}{\textbf{FB15K-237}} &
			\multicolumn{1}{l}{\textbf{NELL-995}} &
			\multicolumn{1}{l}{\textbf{FB15K-237}} &
			\multicolumn{1}{l}{\textbf{NELL-995}} &
			\multicolumn{1}{l}{\textbf{FB15K-237}} &
			\multicolumn{1}{l}{\textbf{NELL-995}} &
			\multicolumn{1}{l}{\textbf{FB15K-237}} &
			\multicolumn{1}{l}{\textbf{NELL-995}} \\
			\cmidrule(lr){2-2} \cmidrule(lr){3-3} \cmidrule(lr){4-4}  \cmidrule(lr){5-5} \cmidrule(lr){6-6} \cmidrule(lr){7-7} \cmidrule(lr){8-8} \cmidrule(lr){9-9}
			\textbf{1p}   & 0.402 & 0.418 & 0.405 & 0.417 & \textbf{0.467} & 0.555          & 0.439          & \textbf{0.587} \\
			\textbf{2p}   & 0.213 & 0.228 & 0.213 & 0.231 & 0.240           & 0.266          & \textbf{0.281} & \textbf{0.305} \\
			\textbf{3p}   & 0.155 & 0.205 & 0.153 & 0.203 & 0.186          & 0.233          & \textbf{0.239} & \textbf{0.326} \\
			\textbf{2i}   & 0.292 & 0.316 & 0.298 & 0.318 & 0.324          & 0.343          & \textbf{0.333} & \textbf{0.371} \\
			\textbf{3i}   & 0.406 & 0.447 & 0.411 & 0.454 & 0.453          & 0.480           & \textbf{0.474} & \textbf{0.531} \\ \cmidrule{1-9}
			\textbf{ip}   & 0.083 & 0.081 & 0.085 & 0.081 & 0.108          & \textbf{0.132} & \textbf{0.110}  & 0.103          \\
			\textbf{pi}   & 0.170  & 0.186 & 0.182 & 0.188 & \textbf{0.205} & \textbf{0.212} & 0.177          & 0.187          \\ \cmidrule{1-9}
			\textbf{Mean} & 0.246 & 0.269 & 0.249 & 0.270  & 0.283          & 0.317          & \textbf{0.293} & \textbf{0.344}
			\\ \bottomrule
		\end{tabular}
	}
	\caption{HITS@3 comparison between \textsc{BiQE}, Q2B and GQE models on FB15K-237, NELL-995 on \textit{conjunctive} queries. Results for Q2B, GQE and GQE-Double were obtained from \cite{Ren2020}. \textbf{ip} and \textbf{pi} are zero shot queries.}
	\label{table:q2b}
\end{table*}

\begin{table}[t!]
	\centering
	\small{
		\begin{tabular}{@{}lcccc}
			\toprule
			& \multicolumn{2}{c}{\textbf{GQE-DistMult-MP}}  & \multicolumn{2}{c}{\textbf{\textsc{BiQE}}} \\
			\cmidrule(lr){2-3} \cmidrule(lr){4-5} 
			& \textbf{MRR}          & \textbf{HITS@10}     &   \textbf{MRR}       & \textbf{HITS@10}               \\
			\cmidrule(lr){2-2} \cmidrule(lr){3-3} \cmidrule(lr){4-4} \cmidrule(lr){5-5}
			
			\textbf{FB15K-237-CQ}             &  0.157            &  0.269            &  \textbf{0.228}         &  \textbf{0.372}    \\ 
			
			\textbf{\textsc{FB15K-237-Paths}} \hspace{-4mm}             &  0.241            &  0.376            &  \textbf{0.473}         &  \textbf{0.602}   \\ 
			\textbf{WN18RR-CQ}            &  0.149            &  0.148           &  \textbf{0.150}         &  \textbf{0.158}  
			\\
			\textbf{\textsc{WN18RR-Paths}}            &  0.349           &  0.400           &  \textbf{0.520}         &  \textbf{0.620}  
			\\
			\bottomrule
		\end{tabular}
	}
	\caption{Comparison of \textsc{BiQE} with best performing version of GQE with Mean Pooling.}
	\label{table:fb15k-237}
	
\end{table}

\begin{table}[]
	\centering
	\ra{1.1}
	\small{
		\begin{tabular}{@{}lcccc@{}}
			\toprule
			& \textbf{GQE-Bilinear} & \textbf{\textsc{TractOR}} & \textbf{MPQE-sum} & \textbf{\textsc{BiQE}}  \\ 
			\cmidrule(lr){2-2} \cmidrule(lr){3-3} \cmidrule(lr){4-4} \cmidrule(lr){5-5}
			\textbf{AUC} & 91.0         & 82.8    & 90.0     & \textbf{96.91} \\
			\textbf{APR} & 91.5         & 86.3    & 90.5     & \textbf{96.69} \\ 
			\bottomrule 
		\end{tabular}
	}
	\caption{Comparing \textsc{BiQE} with GQE, \textsc{TractOr} and MPQE on the \textsc{Bio} dataset. Results for GQE,\textsc{TractOr} and MPQE were obtained from \cite{Hamilton2018, Friedman2020, Daza2020} respectively.}
	\label{table:bio}
\end{table}

\subsection{Results}\label{sec:results}

We compare \textsc{BiQE} to the following reference methods: GQE, GQE-Double, GQE-MP, MPQE, \textsc{TractOR}, and Q2B. GQE, GQE-Double, and Q2B were evaluated using the  FB15K-237 and NELL-995 datasets in \cite{Ren2020}, while GQE, MPQE and \textsc{TractOR} were evaluated on the \textsc{Bio} dataset in \cite{Hamilton2018}, \cite{Daza2020}, and \cite{Friedman2020}, respectively. For a fair comparison we also evaluate \textsc{BiQE} on these datasets.

We present the results on FB15K-237 and NELL-995 in comparison to Q2B and GQE in Table \ref{table:q2b}. For these two datasets, only the first five queries were used for training. The model does not see the last two (\textit{ip} and \textit{pi}) query types during training. The results demonstrate that \textsc{BiQE} is competitive with Q2B and GQE. We would also like to emphasize the observations from \cite{Ren2020}, that simply doubling the embedding dimensions does not necessarily lead to performance improvements. Although \textsc{BiQE} has more parameters compared to GQE and Q2B, simply doubling the parameters (GQE-Double), does not lead to model improvement.

We present results for FB15K-237-CQ, WN18RR-CQ, and the Path datasets in Table \ref{table:fb15k-237}. These datasets study the task of answering queries with multiple missing entities jointly. We do not evaluate Q2B and GQE on the CQ datasets as these approaches cannot be directly used for queries with multiple missing entities. \textsc{BiQE} outperforms the baseline model on both the CQ and Paths datasets. Another interesting observation is that both models are much better at answering path queries compared to DAG queries. Note that the DAG queries were constructed from the same path queries. This shows that DAG queries are more challenging to answer than path queries. We believe this is because DAG queries represent additional constraints which limit the number of correct answers. These results form an additional motivation for the study of probabilistic query answering for DAG-shaped conjunctive queries in KGs. 

Finally we compare \textsc{BiQE} with MPQE, \textsc{TractOR}, and GQE on the \textsc{Bio} dataset in  Table \ref{table:bio}. \textsc{BiQE} improves AUC performance over GQE, the current state of the art, by almost $6$ percentage points.
Table \ref{table:pos} compares model performance for three different positions in the DAG, namely, the tail query, the missing entity at the intersection, and missing entities in the branches of the DAG. Both models are better at predicting missing entities at the intersection of branches compared to the tail entity. This suggests that prediction difficulty depends on the missing entity position in the DAG. 

\begin{figure}[t!]
	\centering
	\includegraphics[width=0.36\textwidth]{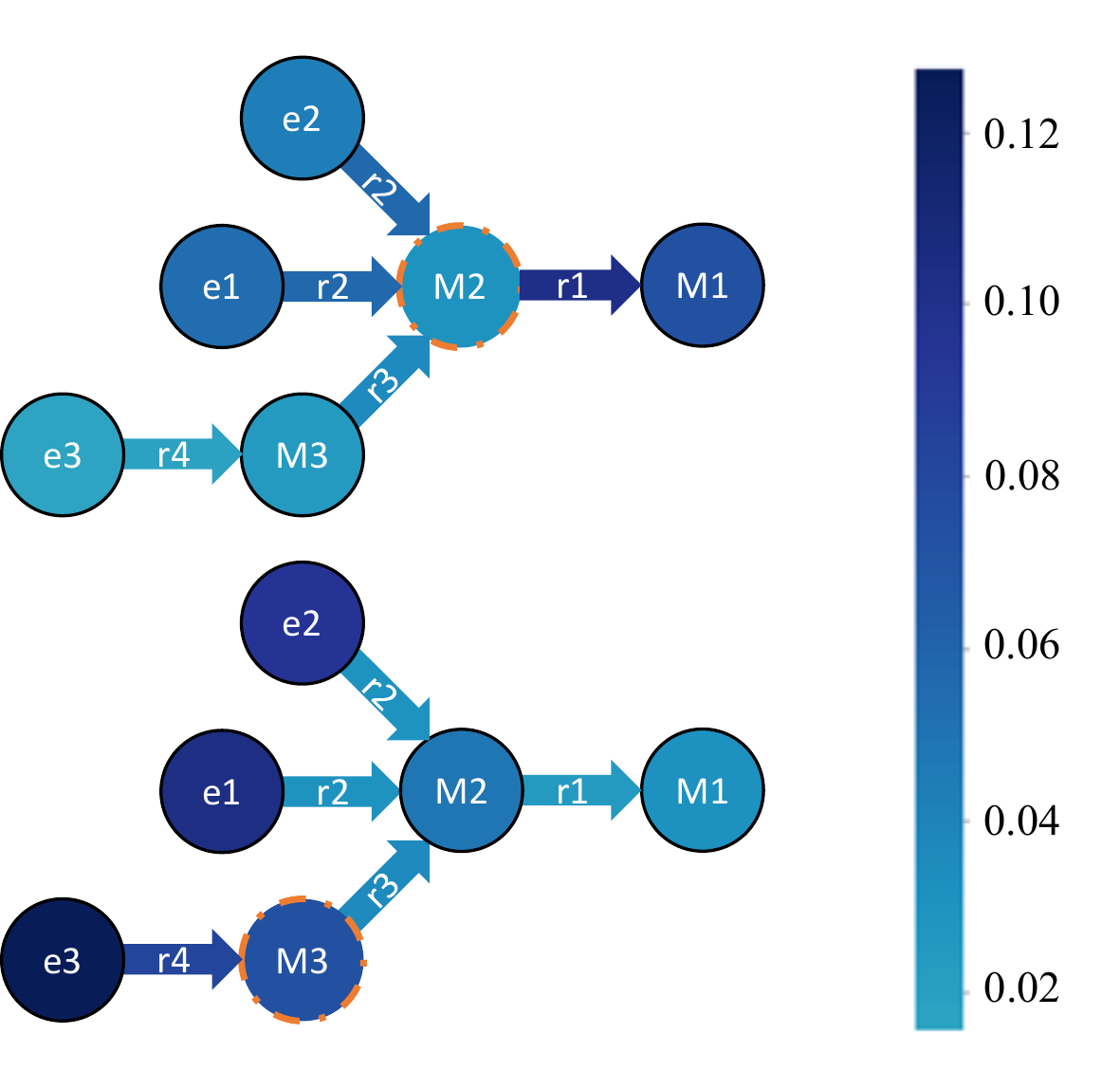}
	\caption{Attention in \textsc{BiQE}: Example of a DAG and attention M2 (top) and M3 (bottom) pay to the other elements in the DAG, where a darker colour equals more attention.}
	\label{fig:attention}
\end{figure}

We believe that the self-attention mechanism is the key factor behind the large improvement. As illustrated in Figure~\ref{fig:embedding}, the GQE model answers queries by traversing the query graph in embedding space in the direction starting from the source entities to the targets. In contrast, the self-attention mechanism allows \textsc{BiQE} to reach missing targets from any arbitrary path or element in the query in the embedding space. The attention matrix and its effect in embedding space illustrates this mechanism (see also Figure \ref{fig:attention}). This suggests that it may be possible to arrive at the missing entities starting from different parts of the query. 

More generally, a graph encoder such as a graph neural network can only aggregate information along the edges of the DAG. Indeed, GQN and Q2B are examples of such graph encoder models. Our experiments show that the real strength of the transformer model is its ability to model all possible interactions between the elements in the query DAG and not just along the existing edges. Figure~\ref{fig:attention} illustrates it inducing a latent graph for every query. The downside of the increased performance is the added cost of computing these interactions through self-attention.

To investigate if the simultaneous prediction and the information flow between all tokens is the crucial feature of \textsc{BiQE}, we conducted additional experiments.

\subsection{Analysis}

For DAG queries involving multiple targets, the \textsc{BiQE} model predicts all the missing entities in one shot. However, it is also possible to predict iteratively where the prediction of the first missing entity is added to the query. We experiment with starting from predicting missing entities closest to the source and ending at the target. Particularly, we would like to confirm that the model benefits from attending to the future query context. To this end, we removed self-attention weights from the path elements occurring to the right (or future) of the current entity that is being predicted \cite{Lawrence2019}. 
For this experiment we used the FB15K-237-Paths dataset and not the FB15K-237-CQ dataset since paths allow for a unique ordering while DAGs allow for partial ordering. This unique ordering allows us to explicitly define notions of past and future query context. Here, the left or the past is the source entity and the future or the right is the missing entity at the end of the terminal relation. The task is to predict all entities other than the source starting from left to right.

As shown in Table \ref{table:analysis}, we indeed see a significant drop in prediction accuracy. This confirms our intuition that it is advantageous to jointly predict the missing entities. It also confirms that the benefits of \textsc{BiQE} stem from the information flow between all tokens and it predicting answers jointly. We confirm these results through qualitative analysis on the test split of the FB15K-237-CQ dataset. We analyse attention pattern for a DAG query in Fig. \ref{fig:attention}. The attention pattern shows that mask 2 (M2) pays attention to mask 1 (M1) and relation 1 (r1) indicating that both M2 and M1 are inferred jointly. In the same DAG, the attention pattern for mask 3 (M3) shows that significant attention is placed on source entities e1 and e2 although they are not the ancestors of M3. GQE, Q2B, and MPQE and its variants cannot aggregate such information from disparate parts of the query. To investigate this phenomenon we compute the attention placed on parts of the query that are not its ancestors or descendants. On average we find that $30.4$ percent of attention is placed on elements that are not the ancestors or descendants of a mask token. This suggests that different elements of the query DAG are useful for predicting the missing entities.

In the camera-ready version we plan to add additional experiments on performance of \textsc{BiQE} on sample of the training set for investigating memorization and run time performance on test set.

\begin{table}[t!]
	\centering
	\ra{1.1}
	\small{
		\begin{tabular}{@{}lccccc@{}}
			\toprule
			& \multicolumn{2}{c}{\textbf{GQE-DistMult-MP}}  & \multicolumn{2}{c}{\textbf{\textsc{BiQE}}} &  \\
			\cmidrule(lr){2-3} \cmidrule(lr){4-5} 
			& \textbf{MRR}          & \textbf{Hits10}     &   \textbf{MRR}       & \textbf{Hits10}             &  \\
			\cmidrule(lr){2-2} \cmidrule(lr){3-3} \cmidrule(lr){4-4} \cmidrule(lr){5-5}
			\textbf{Tail}             &  0.116            &  0.217            &  \textbf{0.205}         &  \textbf{0.319}   &  \\ 
			\textbf{Intersection}            &  0.214            &  0.343           &  \textbf{0.265}         &  \textbf{0.439}  &\\
			\textbf{Branch}            &  0.144            &  0.250           &  \textbf{0.217}         &  \textbf{0.361}  &
			\\ \bottomrule
		\end{tabular}
	}
	\caption{Comparison of \textsc{BiQE} with GQE (Mean Pooling) on the FB15K-237-CQ at various positions in the DAG.}
	\label{table:pos}
	
\end{table}

\section{Related Work}\label{sec:related}
Prior work is primarily focused on the simple link prediction problem using scoring functions operating on entity and relation type embeddings~\cite{Nickel2016}. The KG Question Answering task (KGQA) \cite{Vakulenko2019} aims at answering complex logical queries but focuses only on retrieving answers present in KGs. 
In addition to scoring functions operating on triples \cite{Balazevic2019, Zhang2019, Wang2019Hash, Abboud2020}, other structures such as relational paths \cite{Luo2015, Das2018} and neighborhoods \cite{Schlichktrull2018, Bansal2019, Cai2019} have been used for link prediction. \textsc{BiQE} is related to these methods as it also operates on paths and DAGs. However, unlike prior work, \textsc{BiQE} can answer complex graph queries. 
\begin{table}[]
	\centering
	\ra{1.1}
	\small{
		\begin{tabular}{@{}lll@{}}
			\toprule
			& \textbf{MRR} & \textbf{Hits10} \\
			\cmidrule(lr){2-2} \cmidrule(lr){3-3}
			\textbf{\textsc{BiQE} }&\textbf{0.473}&\textbf{0.602}\\
			\textbf{\textsc{BiQE} (No future context)}& 0.421&0.553\\
			\bottomrule		
		\end{tabular}
	}
	\caption{Predicting one entity at a time (and not jointly) hurts accuracy. Not attending to future context also hurts accuracy. 
	}
	\label{table:analysis}
\end{table}

Apart from the reference models discussed thoroughly in the Experiments Section, models such as the contextual graph attention model, \cite{Mai2019}, also addresses the task of predicting conjunctive queries.\footnote{We are unable to compare our results with them because they used an unreleased, modified Bio dataset.} The contextual graph attention model improves upon GQE by adding an attention mechanism. This is similar to the attention mechanism used in \cite{Ren2020}. \cite{Arakelyan2020} propose Complex Query Decomposition (CQD) that answers complex queries by decomposing them into simple queries and aggregating the results using t-norms. \cite{Fan2019} investigate graph linearization for seq-2-seq models for open domain question answering.

To our knowledge, three recent papers have used transformers for KG completion. \cite{Petroni2019} investigate the use of relational knowledge in pre-trained BERT models for link prediction on open domain KGs. \cite{Yao2019} use pre-trained BERT for KG completion using text. They feed BERT source target entity aliases along with the sentence where the source-target pair co-occurs. However this is problematic due to the large number of sentences needing to be encoded in addition to the noise introduced by distant supervision.
\cite{Wang2019coke} propose Contextualized Knowledge Graph Embedding (CoKE) for answering path queries in knowledge graphs using a Transformer model. Unlike \textsc{BiQE}, however, the method does not answer DAG queries with multiple missing targets, but is limited to path and triple queries with single targets. 
While it may appear that CoKE can be trivially extended to answer DAG queries, this is not the case as there is more than one way to encode DAG queries with a self-attention based model. Indeed, this is the contribution of our work.



\section{Conclusion}\label{sec:conclusion}
We propose a bidirectional self-attention based model for answering conjunctive queries in KGs. We specifically address conjunctive queries whose query graph is a DAG with multiple missing entities. We encode query DAGs as sets of query paths, leveraging a novel positional encoding scheme. Experimentally we showed that \textsc{BiQE} improves upon existing models by a large margin. We showed that the increase in accuracy is due to the bi-directional self-attention mechanism capturing interactions among all elements of a query graph. Furthermore, we  introduced two new benchmarks for studying the problem of conjunctive query answering with multiple missing entities. We limited this work to DAG queries, but we  speculate that \textsc{BiQE} can work well on all kinds of graph queries. This is something We plan to explore in the future.   

\section{Acknowledgements}
We thank the anonymous reviewers for their constructive feedback. We would also like to thank Markus Zopf for fruitful discussions on writing and organizing the manuscript.

\begin{quote}
\begin{small}
\bibliography{biqe}
\end{small}
\end{quote}


\end{document}